\documentclass{article}
\usepackage{graphicx}

\usepackage{amsmath} % \mathbf
\usepackage{amssymb} % \mathbb ...

\usepackage{stmaryrd}% \llbracket
\usepackage{dsfont}  % \mathds
\usepackage{accents} % \accentset

\usepackage[a4paper,margin=2.95cm]{geometry}

\usepackage{authblk}
\title{Zero-shot Learning with Deep Neural Networks for Object Recognition\footnote{This is a preprint of the following chapter: Yannick Le~Cacheux, Hervé~Le Borgne, Michel Crucianu, Zero-shot Learning with Deep Neural Networks for Object Recognition, published in Multi-faceted Deep Learning: Models and Data, edited by Jenny Benois-Pineau, Akka Zemmari, 2021, Springer reproduced with permission of Springer. The final authenticated version is available online at: http://dx.doi.org/}}
\author[1,2]{Le Cacheux, Yannick}
\author[1]{Le Borgne, Hervé}
\author[2]{Crucianu, Michel}
\affil[1]{Université Paris-Saclay, CEA, List, F-91120, Palaiseau, France}
\affil[2]{ CEDRIC -- CNAM, Paris, France}
\date{} 
\begin{document}
\maketitle

\abstract{
Zero-shot learning deals with the ability to recognize objects without any visual training sample. To counterbalance this lack of visual data, each class to recognize is associated with a semantic prototype that reflects the essential features of the object. The general approach is to learn a mapping from visual data to semantic prototypes, then use it at inference to classify visual samples from the class prototypes only. Different settings of this general configuration can be considered depending on the use case of interest, in particular whether one only wants to classify objects that have not been employed to learn the mapping or whether one can use unlabelled visual examples to learn the mapping. This chapter presents a review of the approaches based on deep neural networks to tackle the ZSL problem. We highlight findings that had a large impact on the evolution of this domain and list its current challenges. 
}

\section{Introduction}\label{sec:introduction}
The core problem of supervised learning lies in the ability to generalize the prediction of a model learned on some samples \textit{seen} in the training set to other \textit{unseen} samples in the test set.
A key hypothesis is that the samples of the training set allow a fair estimation of the distribution of the test set, since both result from the same independent and identically distributed random variables. Beyond the practical issues linked to the exhaustiveness of the training samples, such a paradigm is not adequate for all needs, nor reflects the way humans seem to learn and generalize. Despite the fact that, to our knowledge, nobody has seen a real dragon, unicorn or any beast of the classical fantasy, one could easily recognize some of them if met. Actually, from the single textual description of these creatures, and inferring from the knowledge of the real wildlife, there exist many drawings and other visual representations of them in the entertainment industry.

Zero-shot learning (ZSL) addresses the problem of recognizing categories of the test set that are not present in the training set~\cite{larochelle2008zerodata,lampert2009,palatucci_hinton2009zero,farhadi2009describing}. The categories used at training time are called \textit{seen} and those at testing time are \textit{unseen}, and contrary to classical supervised learning, not any sample of unseen categories is available during training. To compensate this lack of information, each category is nevertheless described semantically either with a list of attributes, a set of words or sentences in natural language. The general idea of ZSL is thus to learn some intermediate features from training data, that can be used during the test to map the sample to the unseen classes. These intermediate features can reflect the colors or textures (\textit{fur}, \textit{feathers}, \textit{snow}, \textit{sand}...) or even some part of objects (\textit{paws, claws, eyes, ears, trunk, leaf}...). Since such features are likely to be  present in both seen and unseen categories, and one can expect to infer a discriminative description of more complex concepts from them (e.g.\ some types of animals, trees, flowers...), the problem becomes tractable.

\section{Formalism, Settings and Evaluation}\label{sec:formalism_settings_evaluation}

\subsection{Standard ZSL setting}\label{subsec:zsl_setting}
Formally, let us note the set of seen classes $\mathcal{C^S}$ and that of unseen classes $\mathcal{C^U}$. The set of all classes is $\mathcal{C} = \mathcal{C^S} \cup \mathcal{C^U}$, with $\mathcal{C^S} \cap \mathcal{C^U} =  \emptyset$.

For each class $c \in \mathcal{C}$, \emph{semantic information} is provided. It can consist of binary attributes, such as \textit{``has stripes''}, \textit{``is orange''} and \textit{``has hooves''}.
With this example, the semantic representation of the class \textit{tiger} would be $(1~1~0)^\top$, while the representation of class \textit{zebra} would be $(1~0~1)^\top$.
For a given class $c$, we write its corresponding semantic representation vector $\mathbf{s}_c$; such a vector is also called the \emph{class prototype}. The prototypes of all classes have the same dimension $K$, and represent the same attributes.
More generally, the semantic information does not have to consist of binary attributes, and may not correspond to attributes at all.
More details on the most common types of prototypes are provided in Section~\ref{sec:prototypes}.

For each class $c$, a set of images is available. One can extract a feature vector $\mathbf{x}_i \in \mathbb{R}^D$ from an image, usually using a pre-trained deep neural networksuch as VGG~\cite{simonyan2014very} or ResNet~\cite{he2016deep}. In the latter case, the feature vector of an image corresponds to the internal representation in the network after the last max-pooling layer, before the last fully-connected layer.
It is of course also possible to train a deep network from scratch on the available training images.
In the following, we will refer to an image, a sample of a class or its feature vector with this unique notation $\mathbf{x}$.

During the training phase, the model has only access to the semantic representations of seen classes $\{\mathbf{s}_c\}_{c \in \mathcal{C^S}}$ and to $N$ images belonging to these classes. Hence, the training dataset is $\mathcal{D}^\text{tr} = (\{(\mathbf{x}_n, y_n)\}_{n \in \llbracket 1,N \rrbracket}, \{\mathbf{s}_c\}_{c \in \mathcal{C^S}})$, where $y_n \in \mathcal{C^S}$ is the label of the $n$\textsuperscript{th} training sample.
During the testing phase, the model has access to the semantic representations of unseen classes $\{\mathbf{s}_{c'}\}_{c' \in \mathcal{C^U}}$, and to the $N'$ unlabeled images belonging to unseen classes $\{\mathbf{x}_{n'}\}_{n' \in \llbracket 1,N' \rrbracket}$.
%We will sometimes write $\mathbf{x}^\text{tr}$ or $\mathbf{x}^\text{te}$ to make it explicit whether $\mathbf{x}$ belongs to the training (tr) or testing (te) set, if there is a possible ambiguity.
The objective for the model is to make a prediction $\hat{y}_{n'} \in \mathcal{C^U}$ for each test image $\mathbf{x}_{n'}^{\text{te}}$, assigning it to the most likely unseen class. 

\begin{figure}%[th]
  \centering
  \includegraphics[width=\textwidth]{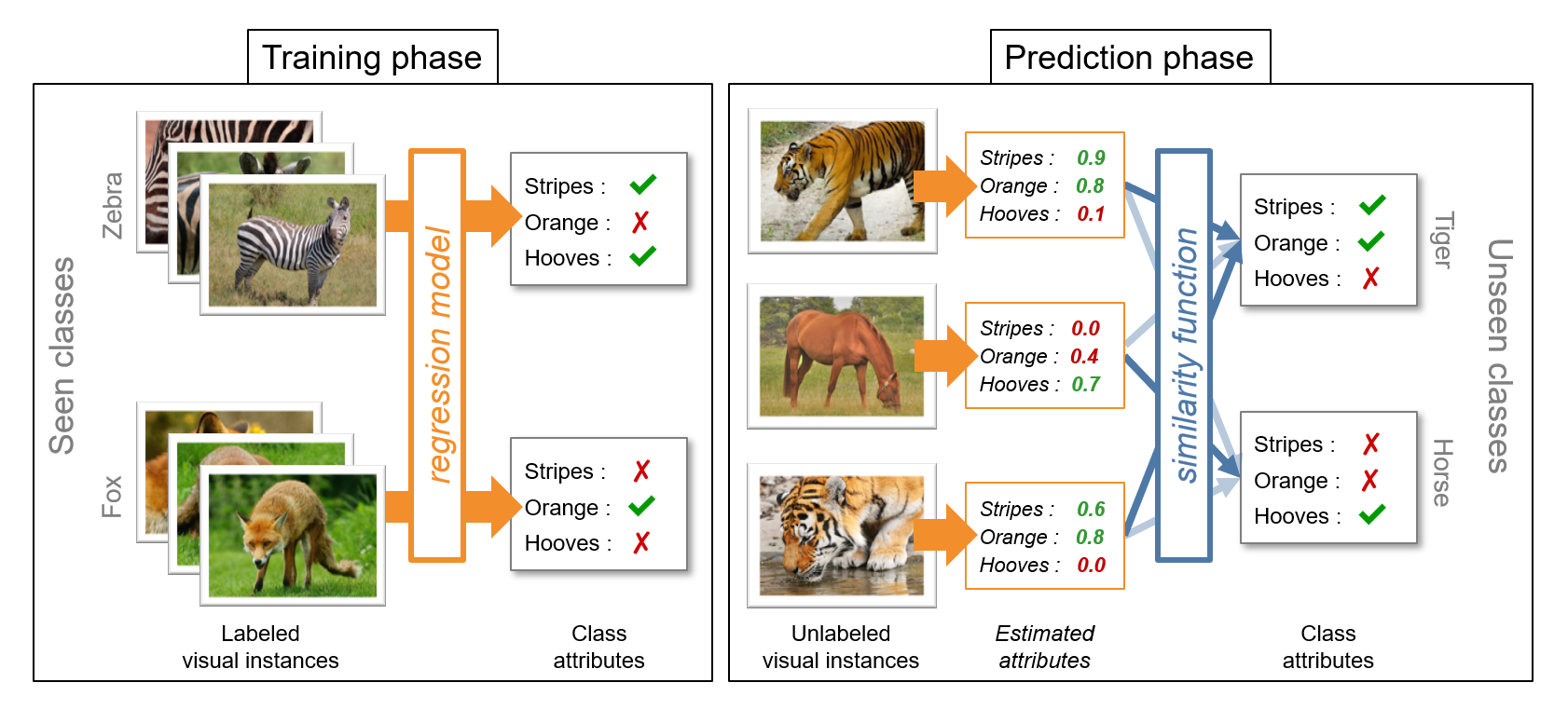}
    \caption{Illustration of a basic ZSL model, with two seen classes \textit{fox} and \textit{zebra}, and two unseen classes \textit{horse} and \textit{tiger}. Each class is represented by a 3-dimensional semantic prototype corresponding to attributes  \textit{``has stripes''}, \textit{``is orange''} and \textit{``has hooves''}. During the training phase, the model learns the relations between the visual features and the attributes using the seen classes. During the evaluation phase, the model estimates the attributes for each test image and predicts the unseen class having the closest prototype.}
    \label{fig:zsl_illustration}
\end{figure}

As a first basic example, a simple ZSL model may consist in simply predicting attributes $\hat{\mathbf{s}}$ corresponding to an image $\mathbf{x}$ such that $\hat{\mathbf{s}}=\mathbf{w}^\top\mathbf{x}$; the parameters $\mathbf{w}$ can be estimated on the training set $\mathcal{D}^\text{tr}$ using a least square loss. To make predictions, we can simply predict the unseen class $c$ whose attributes $\mathbf{s}_c$ are closest to the estimated attributes $\hat{\mathbf{s}}$ as measured by a euclidean distance.
Such a model is illustrated in Fig.~\ref{fig:zsl_illustration}, and will be presented in more details in Section~\ref{subsec:ridge}.

More generally, most ZSL methods in the literature are based on a \emph{compatibility function} $f: \mathbb{R}^D \times \mathbb{R}^K \rightarrow \mathbb{R}$ assigning a ``compatibility'' score $f(\mathbf{x}, \mathbf{s})$ to a pair composed of a visual sample $\mathbf{x} \in \mathbb{R}^D$ and a semantic prototype $\mathbf{s} \in \mathbb{R}^K$, that reflects the likelihood that $\mathbf{x}$ belongs to class $c$ (if $\mathbf{s}$ is $\mathbf{s}_c$). This function may be parameterized by a vector $\mathbf{w}$ or a matrix $\mathbf{W}$, or by a set of parameters $\{\mathbf{w}_i\}_i$, leading to the notation $f_{\mathbf{w}}(\mathbf{x}, \mathbf{s})$ or $f(\mathbf{x}, \mathbf{s}; \{\mathbf{w}_i\}_i)$ in the following. These parameters are generally learned by selecting a suitable loss function $\mathcal{L}$ and minimizing the total training loss $\mathcal{L}_\text{tr}$ over the training dataset $\mathcal{D}^\text{tr}$ with respect to the parameters $\mathbf{w}$:
\begin{equation} \label{eq:loss_framework}
%\frac{1}{N} \sum_{n=1}^{N} \sum_{c \in \mathcal{C^S}} \mathcal{L}(f(\mathbf{x}_n, \mathbf{s}_c), y_n) + \lambda\Omega[f]    % Certaines méthodes (ridge) ne correspondent pas tout à fait à cette formule
\mathcal{L}_\text{tr}(\mathcal{D}^\text{tr}) = 
%\underset{\mathbf{w}}{\text{min}}
\frac{1}{N} \sum_{n=1}^{N} \sum_{c \in \mathcal{C^S}} \mathcal{L}[(\mathbf{x}_n, y_n, \mathbf{s}_c), f_{\mathbf{w}}]
+ \lambda\Omega[f_{\mathbf{w}}]
\end{equation}
where $\Omega[f]$ is a regularization penalty based on $f$ and weighted by $\lambda$. Once %the loss minimized using e.g stochastic gradient descent (SGD), AdaGrad~\cite{duchi2011adagrad}, RMSProp~\cite{hinton2012rmsprop} or Adam~\cite{kingma2014adam}, and 
the model learned, the predicted label $\hat{y}$ of a test image $\mathbf{x}$ can be selected among candidate testing classes based on their semantic representations $\{\mathbf{s}_c\}_{c \in \mathcal{C^U}}$:
\begin{equation} \label{eq:prediction}
\hat{y} = \underset{c \in \mathcal{C^U}}{\text{argmax }} {f_{\mathbf{w}}}(\mathbf{x}, \mathbf{s}_c)
\end{equation}

In the standard setting of ZSL, the only data available during the training phase consists of the class prototypes of the \emph{seen} classes and the corresponding labeled visual samples. The class prototypes of \emph{unseen} classes, as well as the unlabeled instances from these unseen classes, are only provided during the testing phase, after the model was trained. Moreover, the test samples for which we make predictions only belong to these unseen classes. 

\subsection{Alternative ZSL Settings}\label{subsec:alternative_settings}
When class prototypes of both seen \emph{and} unseen classes are available during the training phase, \cite{wang2019survey} considers it as a \emph{class-transductive} setting, as opposed to the standard setting that is \emph{class-inductive}, when unseen class prototypes are only made available after the training of the model is completed.
In a class-transductive setting, the prototypes of unseen classes can for example be leveraged by a generative model, which attempts to synthesize images of objects from unseen classes based on their semantic description (Section~\ref{subsec:generative}). They can also simply be used during training to ensure that the model does not misclassify a sample from a seen class as a sample from an unseen class. An access to this information as early as the training phase may be legitimate for some use-cases but new classes cannot be added as seamlessly as in a class-inductive setting, in which a new class can be introduced by simply providing its semantic representation (without any retraining).

A more permissive setting allows to consider that \emph{unlabeled} instances of unseen classes are available during training.
Such a setting is called \emph{instance-transductive} in~\cite{wang2019survey}, as opposed to the \emph{instance-inductive} setting. These two settings are often simply referred to as respectively \emph{transductive} and \emph{inductive}, even though there is some ambiguity on whether the (instance-)inductive setting designates a class-inductive or a class-transductive setting.
Some methods use approaches which specifically take advantage of the availability of these unlabeled images, for example by extracting additional information on the geometry of the visual manifold~\cite{fu2015transductive}.
Even though models operating in and taking advantage of a transductive setting can often achieve better accuracy than models designed for an inductive setting, one can argue that such a setting is not suitable for many real-life use cases. With a few exceptions~\cite{li2015max}, most transductive approaches consider that the actual (unlabeled) testing instances are available during the training phase, which excludes many practical applications.
Even without this strong assumption, it is not always reasonable to expect to have access to unlabeled samples from many unseen classes during the training phase. One may further argue that this is all the more unrealistic as there is some evidence~\cite{xian2019fvaegan} that labeling even a single instance per class (in a ``one-shot learning'' scenario) can lead to a significant improvement in accuracy over a standard ZSL scenario.
%The default class-inductive, instance-inductive setting is thus the most restrictive setting, but makes the fewest assumptions on the availability of information during the different phases and is therefore the most broadly applicable. 

In some settings, the available information itself can be different from the default setting. For example, in addition to the semantic prototypes, some methods make use of relations between classes defined with a graph~\cite{wang2018knowledgegraphs, kampffmeyer2019rethinking} or a hierarchical structure~\cite{rohrbach2011evaluating}. Others make use of information regarding the environment of the object, for example by detecting surrounding objects~\cite{zablocki2019context} or by computing co-occurence statistics using an additional multilabel dataset~\cite{mensink2014costa}. Other methods consider that instead of a semantic representation per class, a semantic representation \emph{per image} is available, for example in the form of text descriptions~\cite{reed2016learning} or human gaze information~\cite{karessli2017gaze}.

Another classification of ZSL settings is concerned with which classes have to be recognized during the testing phase. Indeed, one may legitimately want to recognize both seen and unseen classes. The setting in which testing instances may belong to both seen and unseen classes is usually called~\emph{generalized zero-shot learning} (GZSL) and has been introduced by~\cite{chao2016generalized}.
Approaches to extend ZSL to GZSL can be divided into roughly two categories: (1)~approaches which explicitly try to identify when a sample does not belong to a seen class, and use either a standard classifier or a ZSL method depending on the result, and (2)~approaches that employ a unified framework for both seen and unseen classes.

In~\cite{socher2013cmt}, the authors explicitly estimate the probability $g_u(\mathbf{x}) = P(y \in \mathcal{C^U} | \mathbf{x})$ that a test instance $\mathbf{x}$ belongs to an unseen class $c \in \mathcal{C^U}$. They first estimate the class-conditional probability density $p(\mathbf{x}|c)$ for all seen classes $c \in \mathcal{C^S}$, by assuming the projections $\hat{\mathbf{s}}(\mathbf{x})$ of visual features in the semantic space %\footnote{\cite{socher2013cmt} actually corresponds to the CMT method from Section~\ref{subsec:ridge}, so $\hat{\mathbf{s}}(\mathbf{x}) = \mathbf{W}_2 \text{tanh}(\mathbf{W}_1\mathbf{x})$ (Equation~\ref{eq:loss_cmt}).}
are normally distributed around the semantic prototype $\mathbf{s}_c$. We can then consider that an instance $\mathbf{x}$ does not belong to a seen class if its class-conditional probability is below a threshold $\gamma$ for all seen classes: 
\begin{equation}
g_u(\mathbf{x}) = \mathds{1}[\forall c \in \mathcal{C^S}, ~p(\mathbf{x}|c) < \gamma]
\end{equation}
%
%Writing $g_s(\mathbf{x}) = 1 - g_u(\mathbf{x})$ the probability that $\mathbf{x}$ belongs to a seen class, it is also possible to use $g_s(\mathbf{x}) \propto \underset{c \in \mathcal{C^S}}{\text{max }} p(\mathbf{x}|c)$ so that estimated probabilities are not binary. The authors of \cite{socher2013cmt} also propose further approaches for estimating $g_u(\mathbf{x})$, for example using unsupervised outlier detection approaches~\cite{kriegel2009loop}.

If one sees the compatibility $f(\mathbf{x}, \mathbf{s}_c)$ as the probability that the label of visual instance $\mathbf{x}$ is $c$,
i.e.\ $P(y=c | \mathbf{x}) \propto f(\mathbf{x}, \mathbf{s}_c)$, %$f(\mathbf{x}, \mathbf{s}_c) \approx P(y=c | \mathbf{x})$,
the compatibilities of seen and unseen classes can be weighted by the estimated probabilities that $\mathbf{x}$ belongs to a seen or unseen class.

% [YLC]: la partie du dessous n'est plus vraiment d'actualité, je l'ai enlevé pour l'instant pour ne pas surcharger trop vite le lecteur avec des détails non indispensables

% so that
% \begin{equation} % On peut virer cette équation, la phrase au dessus est claire
% \hat{y} = \text{argmax}
% \left\{ f(\mathbf{x}, \mathbf{s}_c) g_u(\mathbf{x}) \right\}_{c \in \mathcal{C^U}} \cup
% \left\{ f(\mathbf{x}, \mathbf{s}_c) (1 - g_u(\mathbf{x})) \right\}_{c \in \mathcal{C^S}}
% \end{equation}

% Alternatively, \cite{chao2016generalized} proposed to consider a threshold $\gamma$ (e.g.\ 0.5),
% \begin{equation}
% \hat{y} =
% \begin{cases}
% \underset{c \in \mathcal{C^S}}{\text{argmax }} f(\mathbf{x}, \mathbf{s}_c)
% & \text{if } g_u(\mathbf{x}) \leq \gamma \\
% \underset{c \in \mathcal{C^U}}{\text{argmax }} f(\mathbf{x}, \mathbf{s}_c)
% & \text{if } g_u(\mathbf{x}) > \gamma \\
% \end{cases}
% \end{equation}

% For seen classes $c \in \mathcal{C^S}$, $f(\mathbf{x}, \mathbf{s}_c)$ can be replaced by the output of a standard supervised classifier trained on the seen classes.

Most recent GZSL methods~\cite{verma2018generalized, xian2018fgan, changpinyo2020classifier} adopt a more direct approach: the unweighted compatibility function $f$ is used to directly estimate compatibilities of seen and unseen classes,
so that we simply have
\begin{equation} \label{eq:prediction_gzsl}
\hat{y} = \underset{c \in \mathcal{C^S} \cup \mathcal{C^U}}{\text{argmax }} f(\mathbf{x}, \mathbf{s}_c)
\end{equation}
This approach has the advantage that using a trained ZSL model in a GZSL setting is straightforward, as all there is to do is adding the seen class prototypes to the list of prototypes whose compatibility with $\mathbf{x}$ needs to be evaluated.
However, it has been empirically demonstrated~\cite{chao2016generalized, xian2017goodbadugly} that many ZSL models suffer from a bias towards seen classes.
With the example of Fig.~\ref{fig:zsl_illustration}, many models would thus tend to consider zebras as ``weird'' horses rather than members of a new, unseen class.
To address this problem, a straightforward solution consists in penalizing seen classes to the benefit of unseen classes by decreasing the compatibility of the former by a constant value $\gamma$, similarly to Equation~\ref{eq:calibrated_stacking}. In~\cite{lecacheux2019classical} was put forward a simple method to select a suitable value of $\gamma$ based on a training-validation-testing split specific to GZSL,
which enabled a slight reduction in the accuracy on seen classes to result in a large improvement of the accuracy on unseen classes, thus significantly improving the GZSL score of any model.
%and showed that a slight reduction in the accuracy on seen classes could result in a large improvement of the accuracy on unseen classes, thus significantly improving the GZSL score of any model.

%%%%%%%%%%%%
Other even less restrictive tasks may be considered during the testing or application phase. For instance, one may want a model able to answer that a visual instance matches neither a seen nor an unseen class. Or one may aim to recognize entities that belong to several non-exclusive categories, a setting known as multilabel ZSL~\cite{mensink2014costa, fu2015transductive, lee2018multi}. Other works are interested in the ZSL setting applied to other tasks such as object detection~\cite{bansal2018zsl_detection} or semantic segmentation~\cite{xian2019zsl_segmentation}.

\subsection{ZSL Evaluation}\label{subsec:ZSL_evaluation}
Most of the (G)ZSL works to date address a classification task on mutually exclusive classes, thus the performance is evaluated with a classification rate. The standard accuracy nevertheless computes the score \textit{per sample} (micro-average accuracy).
%, which is not always the best choice for publicly available ZSL datasets~\cite{cub,awa} that are usually well balanced among classes. To better reflect the use case, \cite{xian2017goodbadugly} proposed to compute the score per class (macro-average accuracy) and most recent works adopted this metric.
Although many publicly available ZSL datasets~\cite{cub,awa} have well-balanced classes, other datasets or use cases do not necessarily exhibit this property. \cite{xian2017goodbadugly} therefore proposed to compute the score \textit{per class} (macro-average accuracy) and most recent works adopted this metric.

For GZSL the performance measure is a more subtle issue. Of course, using $y_n \in \mathcal{C^S} \cup \mathcal{C^U}$ for each of the $N$ testing instances, the micro and macro average accuracy can still be employed. However, this does not always provide the full picture regarding the performance of a (G)ZSL model: assuming per class accuracy is used and 80\% of classes are seen classes, a perfect supervised model could achieve 80\% accuracy with absolutely no ZSL abilities. This is all the more important as many GZSL models suffer from a bias towards seen classes, as mentioned previously.

To take the trade-off between seen and unseen classes into account, performance is often measured separately on each type of classes. Chao~\textit{et al.}~\cite{chao2016generalized} defined $\mathcal{A}_{\mathcal{U} \rightarrow \mathcal{U}}$ as the (per class) accuracy evaluated only on test instances of unseen classes when candidate classes are the unseen classes $\mathcal{C^U}$. Also, $\mathcal{A}_{\mathcal{U} \rightarrow \mathcal{C}}$ is the accuracy evaluated on test instances of unseen classes when candidate classes are \emph{all} classes $\mathcal{C}$, seen and unseen. Then $\mathcal{A}_{\mathcal{S} \rightarrow \mathcal{S}}$ and $\mathcal{A}_{\mathcal{S} \rightarrow \mathcal{C}}$ are defined correspondingly. 
Before the GZSL setting, test classes were all unseen classes so the (per-class) accuracy was $\mathcal{A}_{\mathcal{U} \rightarrow \mathcal{U}}$. $\mathcal{A}_{\mathcal{S} \rightarrow \mathcal{S}}$ corresponds to what is measured in a standard supervised learning setting. $\mathcal{A}_{\mathcal{C} \rightarrow \mathcal{C}}$ would correspond to the standard per class accuracy in a GZSL setting.
$\mathcal{A}_{\mathcal{U} \rightarrow \mathcal{C}}$ and $\mathcal{A}_{\mathcal{S} \rightarrow \mathcal{C}}$ respectively measure how well a GZSL model is performing on respectively seen and unseen classes. 
\cite{xian2017goodbadugly} proposes to use the harmonic mean as a trade-off between the two, to penalize models with a high score in one of these two sub-tasks but low performance in the other. This measure is the most commonly employed in the recent GZSL literature \cite{chen2018zero,verma2018generalized,lecacheux2019tripletloss,min2020domain}.
It can be noted that this metric requires to keep some instances from seen classes for the testing phase for a given ZSL benchmark dataset. When this is not convenient, for instance if the number of training samples per class is really small or datasets suffer from biases (see Sec.~\ref{subsec:ZSLdatasets}), sometimes only $\mathcal{A}_{\mathcal{U} \rightarrow \mathcal{C}}$ is evaluated~\cite{hascoet2019generic} in order to still provide some measure of GZSL performance.
Alternatively, \cite{chao2016generalized} introduced \textit{calibrated stacking}, where a weight $\gamma$ is used as a trade-off between favoring $\mathcal{A}_{\mathcal{U} \rightarrow \mathcal{C}}$ (when $\gamma > 0)$ and $\mathcal{A}_{\mathcal{S} \rightarrow \mathcal{C}}$ (when $\gamma < 0$):
\begin{equation} \label{eq:calibrated_stacking}
\hat{y} = \underset{c \in \mathcal{C}}{\text{argmax }} f(\mathbf{x}, \mathbf{s}_c) - \gamma \mathds{1}[c \in \mathcal{C^S}]
\end{equation}
\cite{chao2016generalized} defined the Area Under Seen-Unseen accuracy Curve (AUSUC) as the area under the curve of the plot with $\mathcal{A}_{\mathcal{U} \rightarrow \mathcal{C}}$ on the $x$-axis and $\mathcal{A}_{\mathcal{S} \rightarrow \mathcal{C}}$ on the $y$-axis, when $\gamma$ goes from $-\infty$ to $+\infty$. Similarly to the area under a receiver operating characteristic curve, the AUSUC can be used as a metric to evaluate the performance of a GZSL model.

\subsection{Standard ZSL datasets and evaluation biases}\label{subsec:ZSLdatasets}

\begin{figure}%[bt]
  \centering
  \includegraphics[width=\textwidth]{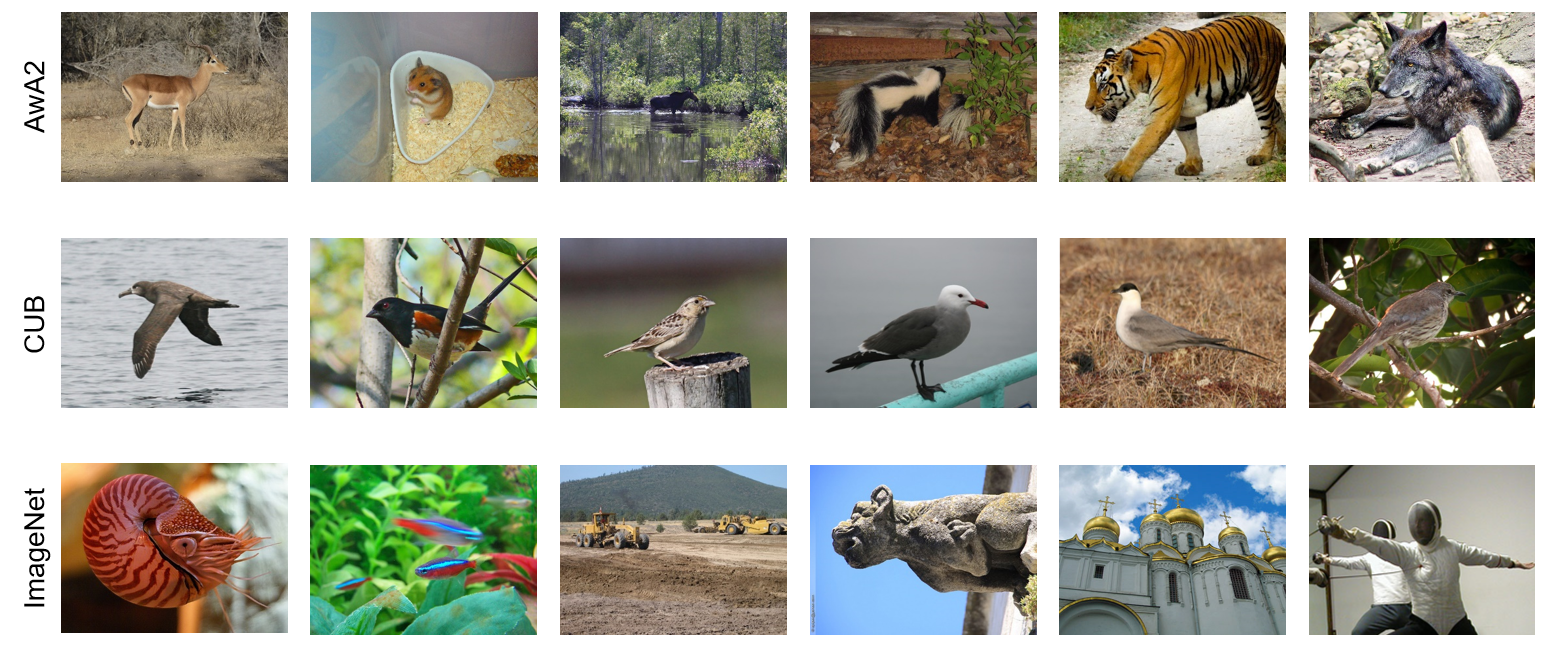}
    \caption{Images from the standard ZSL benchmarks AwA \textit{(top)}, CUB \textit{(middle)} and ImageNet \textit{(bottom)}.}
    \label{fig:zsl_datasets}
\end{figure}

We briefly describe a few datasets commonly used to benchmark ZSL models, provide the rough accuracy obtained on these datasets by mid-2020 %\footnote{This is only to give the reader an idea of the level of performance attained, and may not stay or even be up-to-date.}
and mention a few common biases to avoid when measuring ZSL accuracy on such benchmarks. Some examples of typical images from these datasets are shown in Fig.~\ref{fig:zsl_datasets}. The dataset list is by no means exhaustive, as many other ZSL evaluation datasets can be found in the literature.

\textbf{Animals with Attributes} or \textbf{AwA}~\cite{awa} is one of the first proposed benchmarks for ZSL~\cite{lampert2009}; it has recently been replaced by the very similar \textbf{AwA2}~\cite{xian2018pami} due to copyright issues on some images.
It consists of 37322 images of 50 animal species such as \textit{antelope}, \textit{grizzly bear} or \textit{dolphin}, 10 of which are being used as unseen test classes, the rest being seen training classes. Class prototypes have 85 binary attributes such as \textit{brown}, \textit{stripes}, \textit{hairless} or \textit{claws}.
As mentioned in Sec.~\ref{subsec:zsl_setting}, visual features are typically extracted from images using a deep network such as ResNet pre-trained on a generic dataset like ImageNet.
As evidenced in~\cite{xian2017goodbadugly}, this can induce an important bias on the AwA2 dataset. Indeed, 6 of the 10 unseen test classes are among the 1000 classes of ImageNet used to train the ResNet model; thus, such classes cannot be considered as truly ``unseen''. In~\cite{xian2017goodbadugly} it is therefore proposed to employ a different train / test split, called the \textit{proposed split}, such that no unseen (test) class is present among the 1000 ResNet training classes. This setting has been widely adopted by the ZSL community.
Recent ZSL models in a standard ZSL setting can reach an accuracy of around 71\%~\cite{xian2019fvaegan} on the 10 test classes of this proposed split.

\textbf{Caltech UCSD Birds 200-2011} or \textbf{CUB}~\cite{cub} is referred to as a ``fine-grained'' dataset, as its 200 classes all correspond to bird species (\textit{black footed albatross}, \textit{rusty blackbird}, \textit{eastern towhee}...) and are considered to be fairly similar (Fig.~\ref{fig:zsl_datasets}). Fifty classes are used as unseen testing classes; similarly to AwA2, the standard train / test split has been proposed in \cite{xian2017goodbadugly}. The class prototypes consist of 312 usually continuous attributes with values between 0 and 1. Examples of attributes include \textit{``has crown color blue''}, \textit{``has nape color white''} or \textit{``has bill shape cone''}.
Recent models can reach a ZSL accuracy of around 64\%~\cite{lecacheux2019tripletloss} on the 50 test classes.

The \textbf{ImageNet}~\cite{deng2009imagenet} dataset has also been used as a large-scale ZSL benchmark~\cite{rohrbach2011evaluating}. %,frome2013devise,hascoet2019generic}.
Contrary to AwA or CUB, the usual semantic prototypes do not consist of attributes but rather of word embeddings of the class names -- more details are provided in Sec.~\ref{sec:prototypes}. This dataset contains classes as diverse as \textit{coyote}, \textit{goldfish}, \textit{lipstick} or \textit{speedboat}.
The training classes usually consist of the 1000 classes of the ILSVRC challenge~\cite{ILSVRC15}. In the past, the approximately 20,000 remaining classes were used as unseen test classes. However, \cite{hascoet2019generic} recently showed that this induces a bias, in part due to the fact that unseen classes are often subcategories or supercategories of seen classes. The authors suggested instead to use only a subset of 500 of the total unseen classes such that they do not exhibit this problem.
The best ZSL models in~\cite{hascoet2019generic} can reach an accuracy of around 14\% on these 500 test classes; the fact that this accuracy is significantly lower than on the other two datasets can be attributed to the much larger number of classes, but also to the lower quality of the semantic prototypes (Sec.~\ref{sec:prototypes}).

\section{Methods}\label{sec:methods}
There exist several surveys of the ZSL literature, each with its own classification of existing approaches~\cite{xian2018pami, fu2018survey, wang2019survey}. Here we separate the state of the art into three main categories: \emph{regression methods} (Section~\ref{subsec:ridge}), \emph{ranking methods} (Section~\ref{subsec:triplet_loss}) and \emph{generative methods} (Section~\ref{subsec:generative}). We start by presenting the most simple methods which can be considered as baselines. Sometimes, the methods described below are slightly different from their initial formulation in the original articles, for the sake of brevity and simplicity. We aim at giving a general overview with the strengths, weaknesses and underlying hypotheses of these types of methods, not to dive deep into specific implementation details. As well, we mainly address the GZSL and standard ZSL settings,
%since they are the less restrictive ones.
since they are the most easily applicable to real use-cases.

The \textbf{Direct Attribute Prediction} or \textbf{DAP}~\cite{lampert2009} approach consists in training $K$ standard classifiers which provide the probability $P(a_k|\mathbf{x})$ that attribute $a_k$ is present in visual input $\mathbf{x}$. At test time, we predict the class $c$ which maximizes the probability to have attributes corresponding to its class prototype $\mathbf{s}_c$. Assuming deterministic binary attributes, identical class priors $P(c)$, uniform attribute priors %\footnote{The authors of \cite{lampert2009} also suggest to compute attribute priors $P(\mathbf{a})$ by assuming independent attributes and estimating the probability of each attribute using the training dataset, but indicate that experimental results of both approaches are similar.} 
and independence of attributes, we have:
\begin{equation} \label{eq:dap_prediction_qed}
\underset{c \in \mathcal{C^U}}{\text{argmax }} P(c | \mathbf{x}) =
\underset{c \in \mathcal{C^U}}{\text{argmax }} \prod_{k=1}^K P(a_k = (\mathbf{s}_c)_k | \mathbf{x})
\end{equation}
Similar results may be obtained with continuous attributes by using probability density functions and regressors instead of classifiers.
The \textbf{Indirect Attribute Prediction} or \textbf{IAP} was also proposed in \cite{lampert2009} and is very close to DAP. A notable difference is that it does not require any model training beyond a standard multi-class classifier on seen classes, and in particular does not require any training related to the attributes. As such, it enables to seamlessly convert any pre-trained standard supervised classification model to a ZSL setting provided a semantic representation is available for each seen and unseen class. In Equation~(\ref{eq:dap_prediction_qed}), writing $P(a_k | \mathbf{x})$ as $P(a_k | c)P(c| \mathbf{x})$ and considering that $P(c | \mathbf{x})$ for seen classes can be obtained using any supervised classifier trained on the training dataset on the one hand, and $P(a_k | c)$ is $1$ if $a_k = (\mathbf{s}_c)_k$ and $0$ otherwise on the other hand, we finally have:
\begin{equation}
\hat{y} = \underset{c \in \mathcal{C^U}}{\text{argmax }} \prod_{k=1}^K
\sum_{\underset{(\mathbf{s}_c)_k = (\mathbf{s}_{c'})_k}{c' \in \mathcal{C^S}}} P(c' | \mathbf{x})
\end{equation}
Similarly to IAP, a method based on Convex Semantic Embeddings, or \textbf{ConSE}~\cite{norouzi2013conse}, relies only on standard classifiers and can be used to adapt pre-trained standard models to a ZSL setting without any further training. 
Given a visual sample $\mathbf{x}$, we estimate its semantic representation $\hat{\mathbf{s}}(\mathbf{x}) \in \mathbb{R}^K$ as a convex combination of the semantic prototypes $\mathbf{s}_{\hat{c}}$ of the best predictions $\hat{c}_t(\mathbf{x})$ for $\mathbf{x}$, each prototype being weighted by its classification score.
For a test instance $\mathbf{x}$, we can then simply predict $\hat{y}$ as the class  whose class prototype  is the closest to the estimated semantic representation  as measured with cosine similarity.
We can notice that contrary to DAP and IAP, ConSE does not make any implicit assumption regarding the nature of the class prototypes, and can be used with semantic representations having binary or continuous components. %either binary or continuous semantic representations.
It is also interesting to note that if the convex combination is restricted to one prototype, this method is equivalent to simply finding the best matching seen class to the (unseen) test instance $\mathbf{x}$ and predicting the unseen class whose prototype is closest to the prototype of the best matching seen class.

\subsection{Ridge Regression Approaches}\label{subsec:ridge}
One simple approach to ZSL is to view this task as a regression problem, where we aim to predict continuous attributes from a visual instance. Linear regression is a straightforward baseline. %A very straightforward implementation of this idea consists in learning to predict the attributes from the visual samples with a simple linear regression. 
Given a visual sample $\mathbf{x}$ and the corresponding semantic representation $\mathbf{s}$,
we aim to predict each semantic component $s_k$ of $\mathbf{s}$ as $\hat{s}_k = \mathbf{w}_k^\top \mathbf{x}$, so as to minimize the squared difference between the prediction and the true value $(\hat{s}_k - {s}_k)^2$, $\mathbf{w}_k \in \mathbb{R}^D$ being the parameters of the model.
If we write $\mathbf{W} = (\mathbf{w}_1, \dots, \mathbf{w}_K)^\top \in \mathbb{R}^{K \times D}$, we can directly estimate the entire prototype with $\hat{\mathbf{s}} = \mathbf{W} \mathbf{x}$. We can also directly compare how close $\hat{\mathbf{s}}$ is to $\mathbf{s}$ with $\lVert \hat{\mathbf{s}} - \mathbf{s} \rVert_2^2 = \sum_k (\hat{s}_k - {s}_k)^2$.
% Expliquer pourquoi squared loss et pourquoi régularisation l2 ? (estimateur maximum likelihood, prior sur les paramètres etc)
As with a standard linear regression, we determine the optimal parameters $\mathbf{W}$ by minimizing the error over the training dataset $\mathcal{D}^{\text{tr}}$. 
Let us note $\mathbf{X} = (\mathbf{x}_1, \dots, \mathbf{x}_N)^\top \in \mathbb{R}^{N \times D}$ the matrix whose $N$ lines correspond to the visual features of training samples, and $\mathbf{T} = (\mathbf{t}_1, \dots, \mathbf{t}_N)^\top \in \mathbb{R}^{N \times K}$ that containing the class prototypes associated to each image so that $\mathbf{t}_n = \mathbf{s}_{y_n}$. To simplify notations, we denote $\lVert \cdot \rVert_2$ both the $\ell$2 norm when applied to a vector and the Frobenius norm when applied to a matrix. The loss can then be expressed in matrix form and regularized with an $\ell$2 penalty on the model parameters weighted by a hyperparameter $\lambda$:
\begin{equation} \label{eq:visual_to_semantic_loss}
\frac{1}{N} \lVert \mathbf{XW^\top - \mathbf{T}} \rVert_2^2 + \lambda \lVert \mathbf{W} \rVert_2^2
\end{equation}
Such a loss has a closed-form solution, which directly gives the value of the optimal parameters:
\begin{equation}  \label{eq:visual_to_semantic_solution}
\mathbf{W} = \mathbf{T}^\top\mathbf{X}(\mathbf{X}^\top\mathbf{X} + \lambda N \mathbf{I}_D)^{-1}
\end{equation}
At test time, given an image $\mathbf{x}$ belonging to an unseen class, we estimate its corresponding semantic representation $\hat{\mathbf{s}} = \mathbf{Wx}$ and predict the class with the closest semantic prototype. Note that it's also possible to use other distances or similarity measures such as a cosine similarity during the prediction phase.

% Il manque les "détails" sur la normalisation des vecteurs (qui n'est pas un détail en pratique!)
The \textbf{Embarrassingly Simple approach to Zero-Shot Learning}  \cite{romeraparedes2015eszls}, often abbreviated \textbf{ESZSL}, makes use of a similar idea with a few additional steps. Similarly, the projection $\hat{\mathbf{t}}_n = \mathbf{Wx}_n$ of an image $\mathbf{x}_n$ should be close to the expected semantic representations. This last similarity is nevertheless estimated by a dot product $\hat{\mathbf{t}}_n^\top \mathbf{t}_n$ that should be close to $1$ for the ground truth $t_n=s_{y_n}$ and to $-1$ for the prototypes of other classes. Considering the matrix $\mathbf{Y} \in \{-1, 1\}^{N \times C}$ that is $1$ on line $n$ and column $y_n$ and $-1$ everywhere else, and $\mathbf{S} = (\mathbf{s}_1, \dots, \mathbf{s}_C)^\top \in \mathbb{R}^{|\mathcal{C^S}| \times K}$ the matrix that contains the prototypes of seen classes, the loss to minimize is $\frac{1}{N} \lVert \mathbf{XW}^\top\mathbf{S}^\top - \mathbf{Y} \rVert_2^2$. 
In \cite{romeraparedes2015eszls}, it is further regularized such that visual features projected on the semantic space, $\mathbf{XW}^\top$, have similar norms to allow for fair comparison, and similarly for the semantic prototypes projected on the visual space $\mathbf{W}^\top\mathbf{S}^\top$. Adding an $\ell$2 penalty on the model parameters, we have:
\begin{equation}  \label{eq:eszsl_loss}
\frac{1}{N} \lVert \mathbf{XW}^\top\mathbf{S}^\top - \mathbf{Y} \rVert_2^2
+ \gamma \lVert \mathbf{W}^\top\mathbf{S}^\top \rVert_2^2
+ \frac{\lambda}{N} \lVert \mathbf{XW}^\top \rVert_2^2
+ \gamma \lambda \lVert \mathbf{W} \rVert_2^2
\end{equation}
$\lambda$ and $\gamma$ being hyperparameters controlling the weights of the different regularization terms. The minimization of this expression also leads to a closed-form solution:
\begin{equation}  \label{eq:eszsl_solution}
\mathbf{W} = 
(\mathbf{S}^\top\mathbf{S} + \lambda N \mathbf{I}_K)^{-1} \mathbf{S}^\top\mathbf{Y}^\top\mathbf{X} (\mathbf{X}^\top\mathbf{X} + \gamma N \mathbf{I}_D)^{-1}
\end{equation}

Instead of aiming to predict the class prototypes $\mathbf{s}$ from the visual features $\mathbf{x}$, we can consider \textbf{predicting the visual features} from the class prototypes' features. Each visual dimension is expressed as a linear combination of prototypes such that we can estimate the ``average'' visual representation associated with prototype $\mathbf{s}$ with $\hat{\mathbf{x}} = \mathbf{W}^\top \mathbf{s}$. The distances between the observations and our predictions are then computed in the visual space by minimizing the distance $\lVert \mathbf{x}_n - \hat{\mathbf{x}}_n \rVert^2$ between the sample $\mathbf{x}_n$ and the predicted visual features $\hat{\mathbf{x}}_n = \mathbf{W}^\top \mathbf{s}_{y_n}$ of the corresponding semantic prototype $\mathbf{s}_{y_n}$. The resulting regularized loss $\frac{1}{N} \lVert \mathbf{X - \mathbf{TW}} \rVert_2^2 + \lambda \lVert \mathbf{W} \rVert_2^2$ also has a closed-form solution:
\begin{equation}  \label{eq:semantic_to_visual_solution}
\mathbf{W} = (\mathbf{T}^\top\mathbf{T} + \lambda N \mathbf{I}_K)^{-1} \mathbf{T}^\top\mathbf{X}
\end{equation}
The label of a test image $\mathbf{x}$ is then predicted through a nearest-neighbor search in the visual space. Although this approach is very similar to previous ones, it turns out that projecting semantic objects to the visual space has an advantage. Like other machine learning methods, ZSL methods can be subject to the \emph{hubness problem}~\cite{radovanovic2010hubs}, which describes a situation where certain objects, referred to as \emph{hubs}, are the nearest neighbors of many other objects.
In the case of ZSL, if a semantic prototype is too centrally located, if may be the nearest neighbor of many projections of visual samples into the semantic space even if these samples belong to other classes, thus leading to incorrect predictions and decreasing the performance of the model.
%In the case of ZSL, if a semantic prototype is the nearest neighbor of many projections of visual samples into the semantic space, it can dramatically decrease the performances, since the prototypes are usually much less numerous than images.
When using ridge regression for ZSL, it has been verified experimentally~\cite{lazaridou2015hubnesspollution, shigeto2015hubness} that this situation tends to happen. However, \cite{shigeto2015hubness} shows that this effect is mitigated when projecting from the semantic to the visual space, compared to the opposite situation. It should be noted that the hubness problem does not occur exclusively when using ridge regression, and more complex ZSL methods such as \cite{zhang2017deepembedding} make use of the findings of \cite{shigeto2015hubness}.

The semantic autoencoder (\textbf{SAE})~\cite{kodirov2017sae} approach can be seen as a combination of the two ridge regression projections, from the semantic space to the visual one and the opposite. The idea consists in first \emph{encoding} a visual sample by linearly projecting it onto the semantic space and then \emph{decoding} it by projecting the result into the visual space again. Contrary to the previous proposals, there is no immediate closed-form solution to this problem. However, it can be expressed as a Sylvester equation and a numerical solution can be computed efficiently using the Bartels-Stewart algorithm~\cite{bartels1972sylvester}. During the testing phase, predictions can be made either in the semantic space or in the visual space.

All previous methods project linearly from one modality (visual or semantic) to the other, but they can be adapted to non-linear regression methods, as proposed by \textbf{Cross-Modal Transfer} or \textbf{CMT}~\cite{socher2013cmt}. It consists in a simple fully-connected, 1-hidden layer neural network with hyperbolic tangent non-linearity, which is used to predict semantic prototypes from visual features. Equation~(\ref{eq:visual_to_semantic_loss}) can therefore be re-written as
\begin{equation} \label{eq:loss_cmt}
\frac{1}{N} \sum_n \lVert \mathbf{\mathbf{t}_n} - \mathbf{W}_2 \text{tanh}(\mathbf{W}_1\mathbf{x}_n) \rVert_2^2
\end{equation}
$\mathbf{W}_1 \in \mathbb{R}^{H \times D}$ and $\mathbf{W}_2 \in \mathbb{R}^{K \times H}$ being the parameters of the model, and $H$ the dimension of the hidden layer which is a hyperparameter. Similar or more complex adaptations can easily be made for other methods. The main drawback of such non linear projections compared to the linear methods presented earlier is that there is no general closed-form solution, and iterative numerical algorithms must be used to determine suitable values for the parameters.

\subsection{Triplet-loss Approaches}\label{subsec:triplet_loss}
Triplet loss methods make a more direct use of the compatibility function $f$. The main idea behind these methods is that the compatibility of matching pairs should be much higher than the compatibility of non-matching pairs. More specifically, given a visual sample $\mathbf{x}$ with label $y$, we expect that its compatibility with the corresponding prototype $\mathbf{s}_y$ should be much higher than its compatibility with $\mathbf{s}_{c}$, the prototype of a different class $c \neq y$. How ``much higher'' can be more precisely defined through the introduction of a margin $m$, such that $f(\mathbf{x}, \mathbf{s}_y) \geq m +  f(\mathbf{x}, \mathbf{s}_{c})$. To enforce this constraint, we can penalize triplets $(\mathbf{x}, \mathbf{s}_{y}, \mathbf{s}_{c}), ~c \neq y$, for which this inequality is not true, using the triplet loss
\begin{equation} \label{eq:triplet_penalty}
\mathcal{L}_\text{triplet}(\mathbf{x}, \mathbf{s}_{c}, \mathbf{s}_y; f)
= [m +  f(\mathbf{x}, \mathbf{s}_{c}) - f(\mathbf{x}, \mathbf{s}_y)]_+
\end{equation}
where $[\cdot]_+ = \text{max}(0,\cdot)$. This way, for a given triplet $(\mathbf{x}, \mathbf{s}_{y}, \mathbf{s}_{c}), ~c \neq y$, the loss is 0 if $f(\mathbf{x}, \mathbf{s}_{c})$ is much smaller than $f(\mathbf{x}, \mathbf{s}_y)$, and is all the higher as $f(\mathbf{x}, \mathbf{s}_{c})$ gets close to, or surpasses $f(\mathbf{x}, \mathbf{s}_y)$. In general, it is not possible to derive a solution analytically for methods based on a triplet loss, so we must resort to the use of numerical optimization.

In many triplet loss approaches to ZSL, the compatibility function $f$ is simply defined as a bilinear mapping between the visual and semantic spaces parameterized by a matrix $\mathbf{W} \in \mathbb{R}^{D \times K}$, so that $f(\mathbf{x},\mathbf{s}) = \mathbf{x}^\top\mathbf{Ws}$. This compatibility function is actually the same as with ESZSL, even though the loss function used to learn its parameters $\mathbf{W}$ is different. 
The \textbf{Deep Visual-Semantic Embedding} model or \textbf{DeViSE}~\cite{frome2013devise} is one of the most direct applications of a triplet loss with a linear compatibility function to ZSL: the total loss is simply the sum of the triplet loss over all training triplets $(\mathbf{x}_n, \mathbf{s}_{y_n}, \mathbf{s}_{c}), ~c \neq y$:
\begin{equation}
\mathcal{L}_\text{tr}(\mathcal{D}^{tr})
= \frac{1}{N} \sum_{n=1}^N \sum_{\underset{c \neq y_n}{c \in \mathcal{C^S}}} [m +  f(\mathbf{x}_n, \mathbf{s}_{c}) - f(\mathbf{x}_n, \mathbf{s}_{y_n})]_+
\end{equation}
DeViSE can also be viewed as a direct application of the Weston-Watkins loss~\cite{weston1999multiclasssvm} to ZSL. It can be noted that the link with the generic loss framework in Equation~(\ref{eq:loss_framework}) is this time quite straightforward, as with many triplet loss methods.
Although no explicit regularization $\Omega$ on $f$ is mentioned in the original publication -- even though the authors make use of early stopping in the gradient descent -- it is again straightforward to add an $\ell$2 penalty. 
The \textbf{Structured Joint Embedding} approach, or \textbf{SJE}~\cite{akata2015sje}, is fairly similar to DeViSE. It is inspired by works on structured SVMs \cite{tsochantaridis2004support, tsochantaridis2005structuredsvm}, and makes use of the Cramer-Singer loss~\cite{crammer2001multiclasssvm} for multi-class SVM. Applied to  ZSL, this means that only the class which is violating the triplet-loss constraint the most is taken into account for each sample. In our case, this results in:
\begin{equation} \label{eq:sje_loss}
\mathcal{L}_\text{tr}(\mathcal{D}^\text{tr})
= \frac{1}{N} \sum_{n=1}^N \underset{\underset{c \neq y_n}{c \in \mathcal{C^S}}}{\text{max}} \left( [m +  f(\mathbf{x}_n, \mathbf{s}_{c}) - f(\mathbf{x}_n, \mathbf{s}_{y_n})]_+ \right)
\end{equation}
The \textbf{Attribute Label Embedding} approach or \textbf{ALE}~\cite{akata2013ale, akata2016ale} considers the ZSL task as a ranking problem, where the objective is to rank the correct class $c$ as high as possible on the list of candidate unseen classes. From this perspective, we can consider that SJE only takes into account the top element of the ranking list provided the margin $m$ is close to 0. By contrast, DeViSE penalizes all ranking mistakes: given labeled sample $(\mathbf{x}, y)$, for all classes $c$ mistakenly ranked higher than $y$, we have $f(\mathbf{x}, \mathbf{s}_c) > f(\mathbf{x}, \mathbf{s}_y)$ which contributes to the loss. 
The ALE approach aims to be somewhere in between these two proposals, so that a mistake on the rank when the true class is close to the top of the ranking list weighs more than a mistake when the true class is lower on the list.

Similarly to CMT in the previous section, all triplet-loss models can be extended to the nonlinear case. Such an extension is even more straightforward as this time, having no closed-form solution, all models require the use of numerical optimization. 
One such example of a nonlinear model worth describing due to its historical significance and still fair performance is the \textbf{Synthesized Classifiers} approach, or \textbf{SynC}~\cite{changpinyo2016sync, changpinyo2020classifier}.
Based on a manifold learning framework, it aims to learn \emph{phantom classes} in both the semantic and visual spaces, so that linear classifiers for seen and unseen classes can be synthesized as a combination of such phantom classes. More precisely, the goal is to synthesize linear classifiers $\mathbf{w}_c$ in the visual space such that the compatibility between image $\mathbf{x}$ and class $c$ can be computed with $f(\mathbf{x}, \mathbf{s}_c) = \mathbf{w}_c^\top \mathbf{x}$. The prediction is then $\hat{y} = \underset{c}{\text{argmax }} \mathbf{w}_c^\top \mathbf{x}$. Let us note respectively $\{ \accentset{\ast}{\mathbf{x}}_p\}_{p \in \llbracket 1,P \rrbracket}$ and $\{ \accentset{\ast}{\mathbf{s}}_p\}_{p \in \llbracket 1,P \rrbracket}$ the $P$ phantom classes in the respective visual and semantic spaces. These phantom classes are learned and constitute the parameters of the model\footnote{\label{foot:sync}A number of simplifications were made for the sake of clarity and brevity: in the original article~\cite{changpinyo2016sync},
phantom classes are actually sparse linear combinations of semantic prototypes, $v_{c,p}$ can further use Mahalanobis distance, other losses such as squared hinge loss can be employed instead of the Crammer-Singer loss, Euclidean distances between semantic prototypes can be used instead of a fixed margin in the triplet loss, additional regularization terms and hyperparameters are introduced, and optimization between $\{\accentset{\ast}{\mathbf{x}}_p\}_{p}$ and $\{\accentset{\ast}{\mathbf{s}}_p\}_{p}$ is performed alternatingly.}. 
Each visual classifier is synthesized as a linear combination of visual phantom classes $\mathbf{w}_c = \sum_{p=1}^P v_{c,p} ~\accentset{\ast}{\mathbf{x}}_p$. The value of each coefficient $v_{c,p}$ is set so as to correspond to the conditional probability of observing phantom class $\accentset{\ast}{\mathbf{s}}_p$ in the neighborhood of real class $\mathbf{s}_c$ in the semantic space. Following works on manifold learning~\cite{hinton2003sne, maaten2008tsne}, this can be expressed according to $\mathbf{s}_c$ and $ \accentset{\ast}{\mathbf{s}}_p$. The parameters of the model, i.e.\ the phantom classes $\{(\accentset{\ast}{\mathbf{x}}_p, \accentset{\ast}{\mathbf{s}}_p)\}_{p}$, can be estimated by making use of the Crammer-Singer loss, with adequate regularization to obtain the following objective:
\begin{equation}
\begin{split}
\underset{\{(\accentset{\ast}{\mathbf{x}}_p, \accentset{\ast}{\mathbf{s}}_p)\}_{p}}{\text{minimize }} & 
\frac{1}{N} \sum_{n=1}^N \underset{\underset{c \neq y_n}{c \in \mathcal{C^S}}}{\text{max}} \left( [m +  \mathbf{w}_c^\top \mathbf{x}_n -  \mathbf{w}_{y_n}^\top \mathbf{x}_n]_+ \right) + \\
&  \lambda \sum_{c \in \mathcal{C^S}} \lVert \mathbf{w}_c \rVert^2
+ \gamma \sum_{p=1}^P \lVert \accentset{\ast}{\mathbf{s}}_p \rVert^2
\end{split}
\end{equation}
where $\lambda$ and $\gamma$ are hyperparameters. It is interesting to note that ALE can actually be considered as a special case of SynC, where the classifiers are simply a linear combination of semantic prototypes.

%Other methods relying on a triplet loss include~\cite{chen2018zero}, which uses multiple training objectives and is based on an architecture inspired by adversarial autoencoders~\cite{makhzani2015adversarial}. \cite{li2015max} applies an extension of the Crammer-Singer loss which enables the use of unlabeled samples during training.
Recently, \cite{lecacheux2019tripletloss} showed that modifications to the triplet loss could enable models obtained with this loss to reach (G)ZSL accuracy competitive with generative models (Sec.~\ref{subsec:generative}). Such modifications include a margin that depends on the similarity between $\mathbf{s}_y$ and $\mathbf{s_c}$ in Equation~(\ref{eq:triplet_penalty}) so that confusions between very similar classes are not penalized as much as confusions between dissimilar classes during training, as well as a weighting scheme that makes ``representative'' training samples have more impact than outliers.

\subsection{Generative Approaches}\label{subsec:generative}
Generative methods applied to ZSL aim to produce visual samples belonging to unseen classes based on their semantic description; these samples can then be used to train standard classifiers. Partly for this reason, most generative methods directly produce high-level visual features, as opposed to raw pixels -- another reason being that generating raw images is usually not as effective~\cite{xian2018fgan}.
Generative methods have gained a lot of attention in the last few years: many if not most recent high-visibility ZSL approaches \cite{verma2018generalized, xian2018fgan, xian2019fvaegan} rely on generative models.
This is partly because such approaches have interesting properties, which make them particularly suitable to certain settings such as GZSL.
However, a disadvantage %of these approaches 
is that they can only operate in a class-transductive setting, since the class prototypes of unseen classes are needed to generate samples belonging to these classes; contrary to methods based on regression or explicit compatibility functions, at least some additional training is necessary to integrate novel classes to the model.
We divide generative approaches into two main categories: methods generating a parametric distribution, which consider visual samples follow a standard probability distribution such as a multivariate Gaussian and attempt to estimate its parameters so that visual features can be sampled from this distribution, and non-parametric methods, where visual samples are directly generated by the model.

Methods based on parametric distributions assume that visual features for each class follow a standard parametric distribution. For example, one may consider that for each class $c$, visual features are samples from a multivariate Gaussian with mean $\boldsymbol{\mu}_c \in \mathbb{R}^D$ and covariance $\boldsymbol{\Sigma}_c \in \mathbb{R}^{D \times D}$, such that for samples $\mathbf{x}$ from class $c$ we have $p(\mathbf{x}; \boldsymbol{\mu}_c, \boldsymbol{\Sigma}_c) = \mathcal{N}(\mathbf{x}| \boldsymbol{\mu}_c, \boldsymbol{\Sigma}_c)$. If one can estimate $\boldsymbol{\mu}$ and $\boldsymbol{\Sigma}$ for unseen classes, it is possible to generate samples belonging to these classes. Zero-shot recognition can then be performed by training a standard multi-class classifier on the labeled generated samples.

Alternatively, knowing the (estimated) distribution of samples from unseen classes, one may determine the class of a test visual sample $\mathbf{x}$ using maximum likelihood or similar methods~\cite{verma2017exponential}:
\begin{equation} \label{eq:maximum_likelihood_prediction}
\hat{y} = \underset{c \in \mathcal{C^U}}{\text{argmax }} p(\mathbf{x}; \boldsymbol{\mu}_c, \boldsymbol{\Sigma}_c)
\end{equation}

Other approaches~\cite{xian2018fgan} also propose to further train a ZSL model based on an explicit compatibility function using the generated samples and the corresponding class prototypes, and then perform zero-shot recognition as usually with Equation~(\ref{eq:prediction}).

The \textbf{Generative Framework for Zero-Shot Learning}~\cite{verma2017exponential} or \textbf{GFZSL} assumes that visual features are normally distributed given their class.
The parameters of the distribution $(\boldsymbol{\mu}_c, \boldsymbol{\sigma}^2_c)$ (to simplify, we assume that $\boldsymbol{\Sigma}_c = \text{diag}(\boldsymbol{\sigma}^2_c)$, with $\boldsymbol{\sigma}^2_c \in {\mathbb{R}^{+ D}}$) are easy to obtain for seen classes $c \in \mathcal{C^S}$ using e.g.\ maximum likelihood estimators, but are unknown for unseen classes. Since the only information available about unseen classes consists of class prototypes, one can assume that the parameters $\boldsymbol{\mu}_c$ and $\boldsymbol{\sigma}^2_c$ of class $c$ depend on class prototype $\mathbf{s}_c$. \cite{verma2017exponential} further assumes a linear dependency, such that $\boldsymbol{\mu}_c = \mathbf{W}_\mu^\top ~\mathbf{s}_c$ and $\boldsymbol{\rho}_c = \text{log}(\boldsymbol{\sigma}^2_c) = \mathbf{W}_\sigma^\top ~\mathbf{s}_c$. 
%\begin{equation}
%\boldsymbol{\mu}_c = \mathbf{W}_\mu^\top ~\mathbf{s}_c
%\end{equation}
%\begin{equation} \label{eq:rho_sigma}
%\boldsymbol{\rho}_c = \text{log}(\boldsymbol{\sigma}^2_c) = %\mathbf{W}_\sigma^\top ~\mathbf{s}_c
%\end{equation}
%
The models' parameters $ \mathbf{W}_\mu \in \mathbb{R}^{K \times D}$ and $\mathbf{W}_\sigma \in \mathbb{R}^{K \times D}$ can then be obtained
with ridge regression, using the class distribution parameters $\{(\hat{\boldsymbol{\mu}}_c, \hat{\boldsymbol{\rho}}_c)\}_{c \in \mathcal{C^S}}$ estimated on seen classes as training samples. Similarly to previous approaches, this consists in minimizing $\ell2$-regularized losses, with closed-form solutions. Noting $\mathbf{M} = (\hat{\boldsymbol{\mu}}_1, \dots, \hat{\boldsymbol{\mu}}_{C})^\top \in \mathbb{R}^{C \times D}$ and $\mathbf{R} = (\hat{\boldsymbol{\rho}}_1, \dots, \hat{\boldsymbol{\rho}}_{C})^\top \in \mathbb{R}^{C \times D}$, we have:
\begin{equation}
\mathbf{W}_\mu = (\mathbf{S}^\top\mathbf{S} + \lambda_\mu \mathbf{I}_K)^{-1}\mathbf{S}^\top\mathbf{M}
\end{equation}
\begin{equation}
\mathbf{W}_\sigma = (\mathbf{S}^\top\mathbf{S} + \lambda_\sigma \mathbf{I}_K)^{-1}\mathbf{S}^\top\mathbf{R}
\end{equation}
We can thus predict parameters $(\hat{\boldsymbol{\mu}}_c, \hat{\boldsymbol{\rho}}_c)$ for all unseen classes $c \in \mathcal{C^U}$, and sample visual features of unseen classes accordingly.
Predictions can then be made using either a standard classifier or the estimated distributions themselves. \cite{verma2017exponential} also extends this approach to include more generic distributions belonging to the exponential family and non-linear regressors.

The \textbf{Synthesized Samples for Zero-Shot Learning}~\cite{guo2017synthesizing} or \textbf{SSZSL} method similarly assumes that $p(\mathbf{x}|c)$ is Gaussian, estimates parameters $(\boldsymbol{\mu}, \boldsymbol{\Sigma})$ for seen classes with techniques similar to GFZSL and aims to predict parameters $(\hat{\boldsymbol{\mu}}, \hat{\boldsymbol{\Sigma}})$ for unseen classes. %We can again assume that $\boldsymbol{\Sigma} = \text{diag}(\boldsymbol{\sigma}^2)$.
In a way that reminds the ConSE method, the distributions parameters are estimated using a convex combination of parameters from seen classes $d$, such that $\hat{\boldsymbol{\mu}} = \frac{1}{Z} \sum_{d \in \mathcal{C^S}} w_d \boldsymbol{\mu}_d$ and $\hat{\boldsymbol{\sigma}}^2 = \frac{1}{Z} \sum_{d\in\mathcal{C^S}} w_d \boldsymbol{\sigma}_d^2$, with $Z = \mathbf{1}^\top \mathbf{w} = \sum_d w_d$.
% et techniquement on devrait avoir w_c > 0 non ?
The model therefore has one vector parameter $\mathbf{w}_c \in \mathbb{R}^{|\mathcal{C^S}|}$ to determine per unseen class $c$. 
These last are set such that the semantic prototype $\mathbf{s}_c^\text{te}$ from unseen class $c$ is approximately a convex combination of prototypes from seen classes, i.e.\
$\mathbf{s}_c^\text{te} \simeq \mathbf{S}^\top \mathbf{w}_c / Z_c$,
% ^^^ Référencer une équation de ConSE pour que ce soit plus clair
while preventing classes dissimilar to $\mathbf{s}_c^\text{te}$ from being assigned a large weight. This results in the following loss for unseen class $c$:
\begin{equation} \label{eq:sszsl_loss}
\lVert \mathbf{s}_c^\text{te} - \mathbf{S}^\top \mathbf{w}_c \rVert_2^2 + \lambda \mathbf{w}_c^\top \mathbf{d}_c
\end{equation}
where each element $(\mathbf{d}_c)_i$ of $\mathbf{d}_c$ is a measure of how dissimilar \footnote{In \cite{guo2017synthesizing}, the authors use $(\mathbf{d}_c)_i = \left(\text{exp}\left(-\frac{\lVert \mathbf{s}_c^\text{te} - \mathbf{s}_i^\text{tr} \rVert^2}{\bar{\alpha}^2}\right)\right)^{-1}$ to measure how dissimilar unseen class $c$ is to seen class $i$, where $\bar{\alpha}$ is the mean value of the distances between any two prototypes from seen classes.}
unseen class $c$ is to seen class $i$, and $\lambda$ is a hyperparameter.
Minimizing the second term in Equation~(\ref{eq:sszsl_loss}) naturally leads to assigning smaller weights to classes dissimilar to $c$.
A closed-form solution can then be obtained as:% $\mathbf{w}_c = (\mathbf{SS}^\top)^{-1}(\frac{\lambda}{2}\mathbf{d}_c - \mathbf{Ss}_c^\text{te})$.

\begin{equation}
\mathbf{w}_c = (\mathbf{SS}^\top)^{-1}(\frac{\lambda}{2}\mathbf{d}_c - \mathbf{Ss}_c^\text{te})
\end{equation}

The \textbf{non parametric approaches} do not explicitly make simplifying assumptions about the shape of the distribution of visual features, and use powerful generative methods such as variational auto-encoders (VAEs)~\cite{kingma2013vae} or generative adversarial networks (GANs)~\cite{goodfellow2014generative} to directly synthesize samples. Although these models are in principle able to capture complex data distribution, they can prove to be hard to train~\cite{arjovsky2017towards,plumerault20icpr}, and the resulting images content difficult to control~\cite{plumerault20iclr}.

The \textbf{Synthesized Examples for GZSL} method~\cite{verma2018generalized}, or \textbf{SE-GZSL}, is %an example of an approach 
based on a conditional VAE \cite{sohn2015cvae} architecture.
It consists of two main parts: an \emph{encoder} $\mathcal{E}(\cdot)$ which maps an input $\mathbf{x}$ to an $R$-dimensional internal representation or latent code $\mathbf{z} \in \mathbb{R}^R$, and a \emph{decoder} $\mathcal{D}(\cdot)$ which tries to reconstruct the input $\mathbf{x}$ from the internal representation. An optional third part can be added to the model: a \emph{regressor} $\mathcal{R}(\cdot)$ which estimates the semantic representation ${\mathbf{t}}$ of the visual input $\mathbf{x}$. See Chapter 2 for more details on the VAE architecture. To help the decoder to produce class-dependant reconstructed outputs, the corresponding class prototype $\mathbf{t}_n = \mathbf{s}_{y_n}$ is concatenated to the representation $\mathbf{z}_n$ for input $\mathbf{x}_n$. 

Other approaches such as \cite{mishra2018generative} consider that the encoder outputs a probability distribution, assuming that the true distribution of visual samples is an isotropic Gaussian given the latent representation, i.e. $p(\mathbf{x}|\mathbf{z},\mathbf{t}) = \mathcal{N}(\mathbf{x}|\boldsymbol{\mu}(\mathbf{z}, \mathbf{t}), \sigma^2 \mathbf{I})$. In this case, the output of the decoder should be $\hat{\mathbf{x}} = \boldsymbol{\mu}(\mathbf{z}, \mathbf{t})$, and it can be shown that minimizing $-\text{log}(p(\mathbf{x}|\mathbf{z}, \mathbf{t}))$ is equivalent to minimizing $\lVert \mathbf{x} - \hat{\mathbf{x}} \rVert^2$. Furthermore, in \cite{mishra2018generative}, the class prototype is appended to the visual sample as opposed to the latent code.

The authors of \cite{verma2018generalized} further propose to use the regressor $\mathcal{R}$ to encourage the decoder to generate discriminative visual samples. An example of such components consists in evaluating the quality of predicted attributes from synthesized samples, and takes the form
$\mathcal{L} = -\mathbb{E}_{p(\hat{\mathbf{x}}|\mathbf{z},\mathbf{t})p(\mathbf{z})p(\mathbf{t})}
[\text{log}(p(\mathbf{a}|\hat{\mathbf{x}}))]$.
The regressor itself is trained on both labeled training samples and generated samples, and the parameters of the encoder / decoder and the regressor are optimized alternatingly. 
\textbf{f-GAN} \cite{xian2018fgan} is based on a similar approach, but makes use of conditional GANs~\cite{mirza2014cgan} to generate visual features. It consists of two parts: a \emph{discriminator} $\mathcal{D}$ which tries to distinguish real images from synthesized images, and a \emph{generator} $\mathcal{G}$ which tries to generate images that $\mathcal{D}$ cannot distinguish from real images. Both encoder and decoder are multilayer perceptrons.
The generator is similar to the decoder from the previous approach in that it takes as input a latent code $\mathbf{z} \in \mathbb{R}^R$ and the semantic representation $\mathbf{s}_c$ of a class $c$, and attempts to generate a visual sample $\hat{\mathbf{x}}$ of class $c$: $\mathcal{G}: \mathbb{R}^R \times \mathbb{R}^K \rightarrow \mathbb{R}^D$. The key difference is that the latent code is not the output of an encoder but consists of random Gaussian noise.
The discriminator takes as input a visual sample, either real or generated, of a class $c$ as well as the prototype $\mathbf{s}_c$, and predicts the probability that the visual sample was generated: $\mathcal{D}: \mathbb{R}^D \times \mathbb{R}^K \rightarrow [0,1]$.
$\mathcal{G}$ and $\mathcal{D}$ compete in a two-player minimax game, such that the optimization objective is:
\begin{equation}
\underset{\mathcal{G}}{\text{min }} \underset{\mathcal{D}}{\text{max }}
\mathbb{E}_{p(\mathbf{x},y),p(\mathbf{z})}[\text{log}(\mathcal{D}(\mathbf{x}, \mathbf{s}_y))
+ \text{log}(1 - \mathcal{D}(\mathcal{G}(\mathbf{z}, \mathbf{s}_y), \mathbf{s}_y))]
\end{equation}

The authors of \cite{xian2018fgan} further propose to train an improved Wasserstein GAN \cite{arjovsky2017wasserstein, gulrajani2017improved}, and similarly to \cite{verma2018generalized}, they add another component to the loss to ensure that generated features are discriminative, using a classification loss instead of a regression loss. They call this extended approach f-CLSWGAN.

\section{Semantic Features for Large Scale ZSL}\label{sec:prototypes}

In most of the work on ZSL, the semantic features $\mathbf{s}_c$ were usually assumed to be vectors of attributes such as \textit{``is red''}, \textit{``has stripes''} or \textit{``has wings''}. Such attributes can either be binary %and have values in $\{0,1\}$, or more generally be 
or continuous with values in $[0,1]$. In the latter case, a value of $0.8$ associated with the attribute \textit{``is red''} could mean that the animal or object is mostly red.
But for a large-scale dataset with hundreds or even thousands of classes, or an open dataset where novel classes are expected to appear over time, it is impractical or even impossible to define \emph{a priori} all the useful attributes and manually provide semantic prototypes based on these attributes for all the classes. It is all the more time-consuming as fine-grained datasets may require hundreds of different attributes to reach a satisfactory accuracy~\cite{lecacheux2020webly}.

For large-scale or open datasets it is therefore necessary to identify appropriate sources of semantic information and means to extract this information in order to obtain relevant semantic prototypes. In the case of ImageNet, readily available sources are the word embeddings of the class names and the relations between them according to WordNet~\cite{wordnet1995}, a large lexical database of English that has been developed for many years by human efforts, but is now openly available. Word embeddings are obtained in an unsupervised way and such embeddings of the class names have been employed in ZSL as semantic class representations since~\cite{rohrbach2011evaluating}.
%For large-scale datasets it is therefore common to rely on semantic representations obtained in an unsupervised way instead of using human-defined attributes. Similarly to~\cite{rohrbach2011evaluating}, these representations usually consist in word embeddings of the class names.
The word embedding model is typically trained on a large text corpus, such as Wikipedia, where a neural architecture is assigned the task to predict the context of a given word. 
%M: dommage que Word2Vec ne soit pas détaillé dans un autre chapitre
For instance, the skip-gram objective~\cite{mikolov2013w2v} aims to find word representations containing predictive information about the words occurring in the neighborhood of a given word. Given a sequence $\{w_1, \dots, w_T\}$ of $T$ training words and a context window of size $S$, the goal is to maximize
\begin{equation} \label{eq:skipgram_objective}
\sum_{t=1}^{T}
\sum_{\substack{-S\leq i\leq S \\ i \neq 0}} \text{log }p(w_{t+i}|w_t)
\end{equation}
Although deep neural architectures could be used for this task, it is much more common to use a shallow network with a single hidden layer. In this case, each unique word $w$ is associated with an ``input'' vector $\mathbf{v}_{w}$ and an ``output'' vector $\mathbf{v}_{w}'$, and $p(w_i|w_t)$ is computed such that
\begin{equation}
p(w_i|w_t) = \frac{\text{exp}(\mathbf{v}_{w_i}'^\top\mathbf{v}_{w_t})}{\sum_{w} \text{exp}(\mathbf{v}_{w}'^\top\mathbf{v}_{w_t})}
\end{equation}
The internal representation corresponding to the hidden layer, i.e.\ the input vector representation $\mathbf{v}_{w}$, can then be used as the word embedding. Other approaches such as~\cite{bojanowski2017fasttext} or \cite{pennington2014glove} have also been proposed.

%M: intérêt pas clair dans ce contexte
%Such embeddings have interesting properties. For instance, as illustrated in~\cite{mikolov2013w2v}, the relation (i.e.\ the difference) between the embeddings of the words \textit{``Russia''} and \textit{``Moscow''} is very similar to the relation between \textit{``France''} and \textit{``Paris''}.

Semantic information regarding ImageNet classes can also be provided by WordNet subsumption relationships between the classes (or IS-A relations). They were obtained by several methods, e.g.\ with graph convolutional neural networks, and employed for ZSL on ImageNet with state-of-the-art results~\cite{wang2018knowledgegraphs, kampffmeyer2019rethinking}. However, as shown in~\cite{hascoet2019generic}, these results are biased by the fact that in the traditional ImageNet ZSL split between seen and unseen classes, the unseen classes are often subcategories or supercategories of seen classes. When the ZSL split is modified so as to remove this bias, the WordNet graph-based embeddings lead to an accuracy of about 14\% according to~\cite{hascoet2019generic}.
%The hypothesis behind the use of these embeddings as class prototypes is that these embeddings encompass visual information in addition to this linguistic information.
However, although using word or graph embeddings can reduce the additional human effort required to obtain class prototypes to virtually zero -- pre-trained word embeddings can easily be found online -- there still exists a large performance gap between the use of such embeddings and manually crafted attributes~\cite{lecacheux2020webly}. 
Such a difference may be due in part to the text corpora used to train the word embeddings.%: standard corpora.

The use of complementary sources to produce semantic class representations for ZSL relies on the assumption that the information these sources provide reflects visual similarity relations between classes. However, the word embeddings are typically developed from generic text corpora, like Wikipedia, that do not focus on the visual aspect. Also, the subsumption relationships issued from WordNet and supporting the graph-based class embeddings represent hierarchical conceptual relations. In both cases, the sought-after visual relations are at best represented in a very indirect and incomplete way.

To address this limitation and include more visual information to build the semantic prototypes, it was recently proposed to employ text corpora with a more visual nature, by constructing such a corpus from Flickr tags~\cite{lecacheux2020webly}. Following~\cite{popescu2011social}, the authors of \cite{lecacheux2020webly} further suggested to address the problem of \textit{bulk tagging}~\cite{o2013modeling} -- users attributing the exact same tags to numerous photos -- by ensuring that a tuple of words $(w_i, w_j)$ can only appear once for each user during training, thus preventing a single user from having a disproportionate weight on the final embedding. 
Also, \cite{lecacheux2020using} suggested to exploit the sentence descriptions of WordNet concepts, in addition to the class name embedding, to produce semantic representations better reflecting visual relations.
Any of these two proposals allow to reach an accuracy between 17.2 and 17.8 on the 500 test classes of the ImageNet ZSL benchmark with the linear model from the semantic to the visual space (Section~\ref{subsec:ridge}), compared to 14.4 with semantic prototypes based on standard embeddings.

\section{Conclusion and current challenges}\label{sec:conclusion}

Zero-shot learning addresses the problem of recognizing categories that are missing from the training set. ZSL has grown from an endeavor of some machine learning and computer vision researchers to find approaches that come closer to how humans learn to identify object classes. It now aims to become a radical answer to the concern that the amount of labeled data grows much slower than the volume of data in general, so supervised learning alone cannot produce satisfactory solutions for many real-world applications.

Key to the possibility of recognizing previously unseen categories is the availability for all categories, both seen and unseen, of more than just conventional labels. For each category we should have complementary information (or features) reflecting the characteristics of the modality used for recognition (visual if recognition is directed to images). The relation between these features and the target modality can thus be learned from the seen categories and then employed for recognizing unseen categories. 

Most of the work on ZSL took advantage of the existence of some small or medium-size datasets for which the complementary information, under the form of attributes, was devised and manually provided to support the development of ZSL methods. However, in general applications one has to deal with large and even open sets of categories, so other approaches should be found for identifying associated sources of complementary information and exploiting them in ZSL.

For the large ImageNet dataset several readily available complementary sources were found, including word embeddings of class names, WordNet-based concept hierarchies including the classes as nodes, and short textual definitions from WordNet. While this allowed to extend ZSL methods to such large-scale datasets, the state-of-the-art accuracy obtained on the unseen categories of ImageNet is yet disappointing. This is because the information provided by these sources reflects mostly the conceptual relations and not so much the visual characteristics of the categories. To go beyond this level of performance we consider that two important steps should be taken. First, it is necessary to assemble large corpora including rather detailed textual descriptions of the visual characteristics of a large number of object categories. Partial corpora do exist in various domains (e.g.\ flora descriptions) and different languages. Second, zero-shot recognition should rely on a deeper, compositional analysis  (e.g.\ \cite{purushwalkam2019task}) 
of an image as well as on visual reasoning.

\bibliographystyle{alpha}
\bibliography{biblio}

\end{document}